\documentclass[journal]{IEEEtran}
\usepackage{amsmath,amssymb,amsfonts}
\usepackage{booktabs,multirow}
\usepackage{graphicx}
\usepackage{xcolor}
\usepackage{url}
\newcommand{\vect}[1]{\mathbf{#1}}

\begin{document}
\title{Selective Event Evidence and Persistent Temporal Priors for Event-Based Point Tracking}
\author{Xiaokai~Song,~Zhang~Li,~Leqi~Liu,\\
        Yibin~Ye,~Kun~Wang,~and~Xichao~Teng%
\thanks{This work was supported in part by the National Natural Science Foundation of China under Grant 12472189, the Science and Technology Innovation Program of Hunan Province under Grant 2022RC1196, and the National Key Laboratory of Advanced Propulsion Technology Fund under Grant No. 25-GC-ZZKY-021.}
\thanks{Xiaokai Song, Zhang Li, Yibin Ye, Leqi Liu, and Xichao Teng are with the College of Aerospace Science and Engineering, National University of Defense Technology, Changsha 410000, China, and with the Hunan Provincial Key Laboratory of Image Measurement and Vision Navigation, National University of Defense Technology, Changsha 410000, China.}
\thanks{Kun Wang is with the Videogrammetry Innovation Center, Hunan Institute of Advanced Technology.}
\thanks{Corresponding author: Zhang Li (e-mail: lizhang08@nudt.edu.cn)}%
}
\maketitle

\begin{abstract}
Event cameras enable tracking under fast motion and challenging illumination. However, stable long-term point tracking remains challenging: event activity is sparse and varies with scene content and motion, complicating correspondence, while trajectory estimates can drift over long sequences. Together, these challenges motivate a tracker that selectively uses informative event evidence and retains trajectory constraints across temporal windows. We propose Selective Event Evidence and Persistent Temporal Priors (SEPT), an event-only framework for tracking any point (TAP) over long sequences. SEPT introduces two complementary modules: Task-Adaptive Event Evidence Modulation (TEEM) and Persistent Trajectory Prior (PTP). Using only tracking supervision, TEEM learns a bounded reliability field that weights the event tensor across space, time, and channels before feature extraction. PTP carries geometric information across overlapping windows by marginalizing a fixed local quadratic trajectory model with a Schur complement. Its visibility-aware correction refines overlap coordinates, while a clipped terminal velocity initializes coordinates at newly exposed timestamps. SEPT builds on the ETAP architecture. SEPT is trained exclusively on synthetic EventKubric data and evaluated on five datasets across TAP and feature tracking. Results show that SEPT outperforms the retrained ETAP baseline in average Jaccard for TAP and in feature age and expected feature age for feature tracking; it also surpasses several methods combining event and frame inputs on EC and EDS.
The source code is publicly available at \url{https://github.com/sssxxxkkkk/SEPT}.
\end{abstract}

\begin{IEEEkeywords}
Event cameras, tracking any point, point tracking, event representation, temporal prior, Schur complement.
\end{IEEEkeywords}

\begin{figure*}[!t]
\centering
\includegraphics[width=0.9\textwidth]{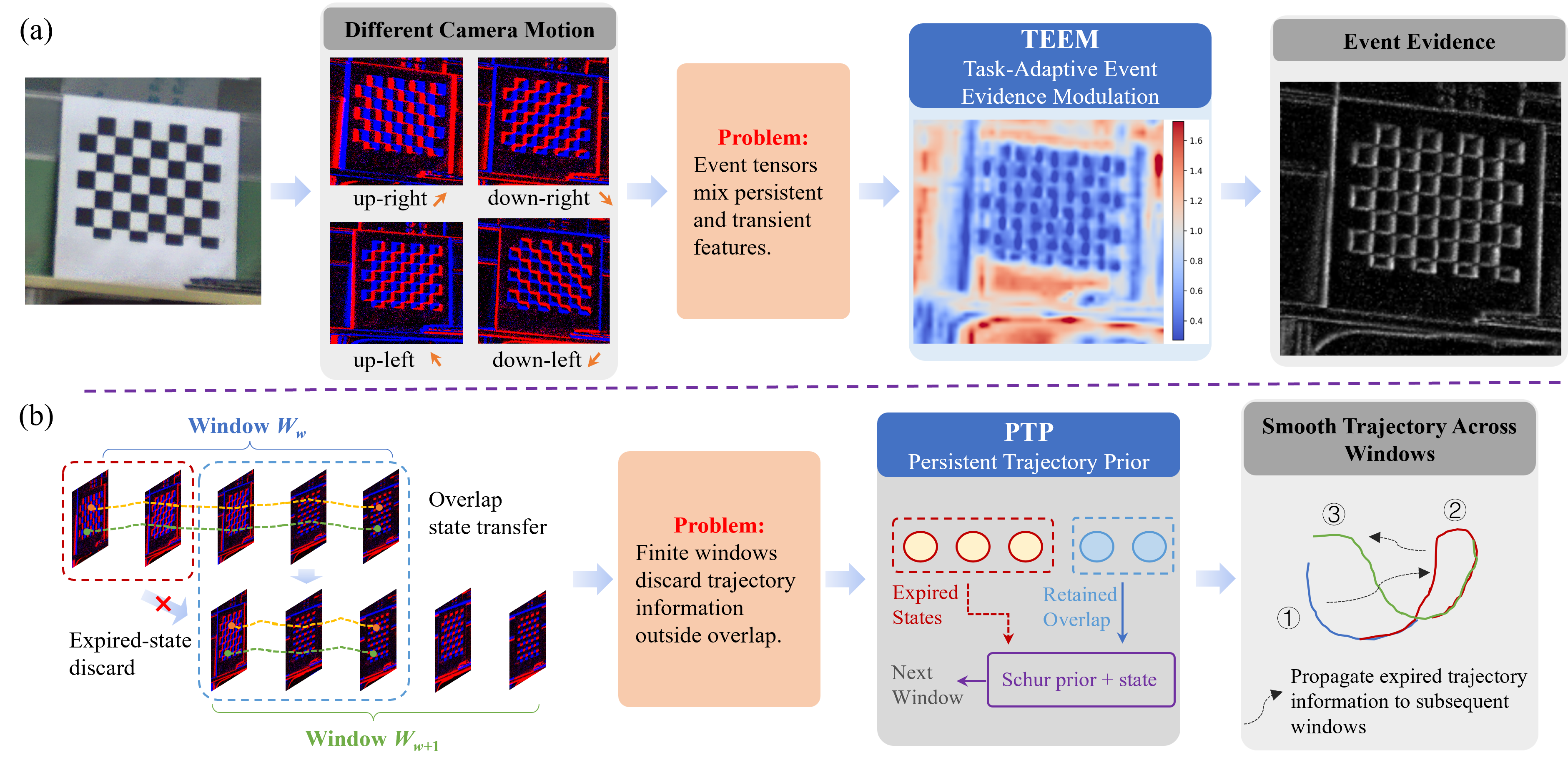}
\caption{Bottlenecks and corresponding solutions in event-based point tracking. (a) Sparse, nonuniform event activity yields competing local matches; TEEM learns task-adaptive event modulation before feature extraction. (b) As the tracking window advances, historical constraints are discarded and trajectories may drift; PTP carries a Schur-style trajectory prior into the next window.}
\label{fig:motivation}
\end{figure*}

\section{Introduction}
Tracking any point (TAP) estimates the trajectory and visibility of an arbitrary query throughout a sequence \cite{doersch2022tap}. Such tracks provide a geometric substrate for tasks including augmented reality, robot manipulation, visual odometry, and dynamic-scene analysis \cite{doersch2022tap,li2024taptrv2,vecerik2024robotap,chen2024leapvo}. The field has progressed from local feature correspondence to joint reasoning over many queries and time steps: TAP-Vid~\cite{doersch2022tap} formalized a broad evaluation protocol, while PIPs~\cite{harley2022pips}, TAPIR~\cite{doersch2023tapir}, and CoTracker~\cite{karaev2024cotracker,karaev2024cotracker3} refine trajectories using temporal context, matching, or collective motion cues. Nevertheless, frame-based observations can become unreliable when exposure causes blur, illumination saturates the sensor, or inter-frame motion is large \cite{harley2022pips,doersch2023tapir,lemoing2024dot}.

Event cameras provide a complementary sensing modality. Instead of capturing complete images at a fixed rate, these sensors asynchronously emit brightness-change events, providing high temporal resolution, high dynamic range, and minimal motion blur \cite{lichtsteiner2008dvs,gallego2022survey}. These properties have enabled event-based feature tracking and continuous trajectory estimation in regimes that are difficult for standard cameras \cite{gehrig2019eklt,mueggler2015continuous,messikommer2023ddft,li2024event3d}. They do not make correspondence automatic, however. Event activity is sparse and depends jointly on scene gradients, relative motion, contrast sensitivity, and the temporal interval used to form a dense representation. Thus, the event pattern produced by one surface can change substantially across motions and time scales.

This observation motivates event-based TAP, which seeks the broad query coverage of modern TAP while retaining the sensing advantages of events. ETAP~\cite{hamann2025etap} demonstrates that event-only tracking can be built on motion-robust correlations and feature alignment, while MATE~\cite{han2025mate} adds explicit motion guidance for variable dynamics. A remaining issue is that an accumulated event tensor mixes persistent correspondence cues with responses that may be weak, transient, or poorly localized. We therefore investigate whether tracking supervision alone can guide a tracker to emphasize the event responses most useful for correspondence. This motivates our first component, Task-Adaptive Event Evidence Modulation (TEEM).

Long-term tracking introduces a second challenge. Events arrive continuously, whereas modern joint trackers must use bounded temporal windows to control the cost of feature encoding, correlation construction, and transformer refinement~\cite{harley2022pips,karaev2024cotracker}. Sliding windows with iterative updates are a practical solution, and overlapping windows allow a tracker to carry forward current coordinate estimates~\cite{karaev2024cotracker,wang2024toward}. Yet a window transition still truncates the optimization history. States that leave the active window are discarded, even though they constrain the trajectory in the overlap and may remain useful for subsequent refinement. Simply copying an overlap estimate provides an initialization, but it does not preserve a structured account of the information contributed by the discarded states. Recurrently copying high-dimensional latent features is also not necessarily reliable, because those features are continually updated in a window-specific correlation context. This motivates our second component, the Persistent Trajectory Prior (PTP), which retains trajectory information without requiring unbounded back-propagation or a large learned memory.

As illustrated in Fig.~\ref{fig:motivation}, Selective Event Evidence and Persistent Temporal Priors (SEPT) addresses these two issues with an event-only extension of the ETAP~\cite{hamann2025etap} tracking architecture. TEEM predicts a bounded reliability field over the input event tensor. It combines spatiotemporal summary statistics---signed response, response magnitude, and temporal variation---with channel-wise statistics. The resulting weight is applied before event feature extraction. TEEM is trained using only point-tracking supervision: it receives no contour labels, segmentation masks, or event-importance annotations. Its zero-initialized final projections make TEEM an identity mapping at the start of training, allowing it to learn selective modulation without perturbing the initial event representation.

PTP preserves trajectory-level information across overlapping windows. For a completed window, we construct a small quadratic model over coordinate perturbations with observation and second-order temporal-smoothness terms. We partition its states into those that leave the window and those that remain in the overlap, and analytically eliminate the former using a Schur complement. The resulting precision matrix provides a compact prior on the retained coordinates. In the next window, stored overlap coordinates and visibility logits initialize the recurrent state, while a clipped terminal velocity initializes coordinates at newly exposed timestamps. The prior then applies a visibility-aware correction at every coordinate-refinement iteration. This is a Schur-style prior rather than a full bundle-adjustment marginalization: it is derived from a fixed trajectory model, not from a data-dependent linearization of visual residuals. 

The two components operate at complementary information bottlenecks. TEEM improves the quality of observation evidence before it reaches the encoder and correlation module. PTP improves the persistence of solved geometric information after a temporal window shifts. This separation is useful in practice. Better event evidence can produce a more reliable overlap trajectory, while a structured temporal prior can reduce boundary discontinuities in regions with sparse event evidence. Neither mechanism replaces the other.

Our contributions are threefold:
\begin{itemize}
\item We introduce TEEM, a lightweight, identity-initialized module that learns coupled spatiotemporal and channel-wise weights for event evidence from downstream tracking supervision.
\item We introduce PTP, a parameter-free fixed-lag trajectory prior that preserves the influence of expired window states through a Schur complement and applies a visibility-aware, preconditioned coordinate correction during iterative tracking.
\item We evaluate SEPT under an event-only training protocol on five datasets for TAP and feature tracking. We analyze the contributions of TEEM and PTP and the sensitivity to training-data allocation.
\end{itemize}

The remainder of the paper is organized as follows. Section II reviews related work, Section III presents SEPT, Section IV reports experimental results, Section V discusses limitations and future directions, and Section VI concludes the paper.

\section{Related Work}
\subsection{Event-Based Feature Point Tracking}
Early event-based tracking focuses on a sparse set of salient features. One family of methods groups nearby events in the $(x,y,t)$ domain and estimates a feature state from each cluster. For example, eCDT~\cite{hu2022ecdt} clusters adjacent events with the same polarity and matches cluster descriptors to extend feature tracks directly from the event stream. By contrast, HOTS~\cite{lagorce2017hots} encodes the recency of activity at each pixel, yielding local spatiotemporal patterns for comparison or clustering. A related point-set formulation builds an edge or event template around a detected feature and aligns incoming events through ICP registration~\cite{kueng2016lowlatency,zhu2017probabilistic}. EKLT~\cite{gehrig2019eklt} combines event-driven updates with frame-based feature detection, using the stable photometric information in frames and the high temporal resolution of events. RATE~\cite{ikura2024rate} combines asynchronous feature initialization with track management for real-time operation. Learned methods instead use local features and association functions, as in DDFT~\cite{messikommer2023ddft}, EventPoint~\cite{huang2023eventpoint}, and 3D Feature Tracking via Event Camera~\cite{li2024event3d}. At the object level, DANet~\cite{fu2023distractor} uses event motion and contour cues, while CrossEI~\cite{chen2025crossei} fuses event and frame cues. These target-tracking formulations differ from arbitrary-query point tracking.

Dense optical-flow estimation provides a complementary representation by predicting a displacement field over the image or active pixels \cite{zhu2019evflownet,gehrig2021e2vid}. It is useful for short-term motion, but it does not by itself maintain the identity and visibility of an arbitrary query. Moreover, flow remains susceptible to the aperture problem and occlusion, and its estimate depends on the temporal interval used to aggregate events. Errors can therefore accumulate when local flow estimates are chained into long trajectories. These limitations motivate a tracker that explicitly maintains query-conditioned trajectories rather than only dense instantaneous motion.

\subsection{Tracking Any Point}
TAP extends sparse feature tracking to arbitrary query points, including those in weakly textured or non-corner regions and those that become temporarily occluded, by estimating their trajectories and visibility. TAP-Vid~\cite{doersch2022tap} established a unified benchmark and evaluation protocol for this setting. Frame-based methods have subsequently evolved from local matching to iterative trajectory refinement and joint reasoning across time and tracks. PIPs~\cite{harley2022pips} refines point trajectories over a sequence rather than treating frame pairs independently, PIPs++~\cite{zheng2023pipsplusplus} extends the temporal receptive field and uses multiple appearance templates for long-term tracking, and TAPIR~\cite{doersch2023tapir} adds per-frame initialization and temporal refinement. CoTracker~\cite{karaev2024cotracker,karaev2024cotracker3} further exploits interactions among many trajectories and uses causal, overlapping windows to support efficient online inference.

Applying TAP to event cameras is nontrivial. Unlike frame features, events provide only brightness changes; their spatial density varies with texture and motion, and the same surface can generate different event patterns under different velocities or directions. ETAP~\cite{hamann2025etap} established event-only TAP with motion-robust correlation features and feature alignment. MATE~\cite{han2025mate} adds motion guidance and variable-motion-aware processing to reduce ambiguities in sparse event observations. These works establish the feasibility of event-based TAP, but responses within an accumulated event tensor differ in how reliably they support point correspondence. TEEM addresses this issue before feature extraction by learning a bounded, task-driven weighting over the event representation, without requiring labels for event types or manually designed importance rules.

\begin{figure*}[!t]
\centering
\includegraphics[width=\textwidth]{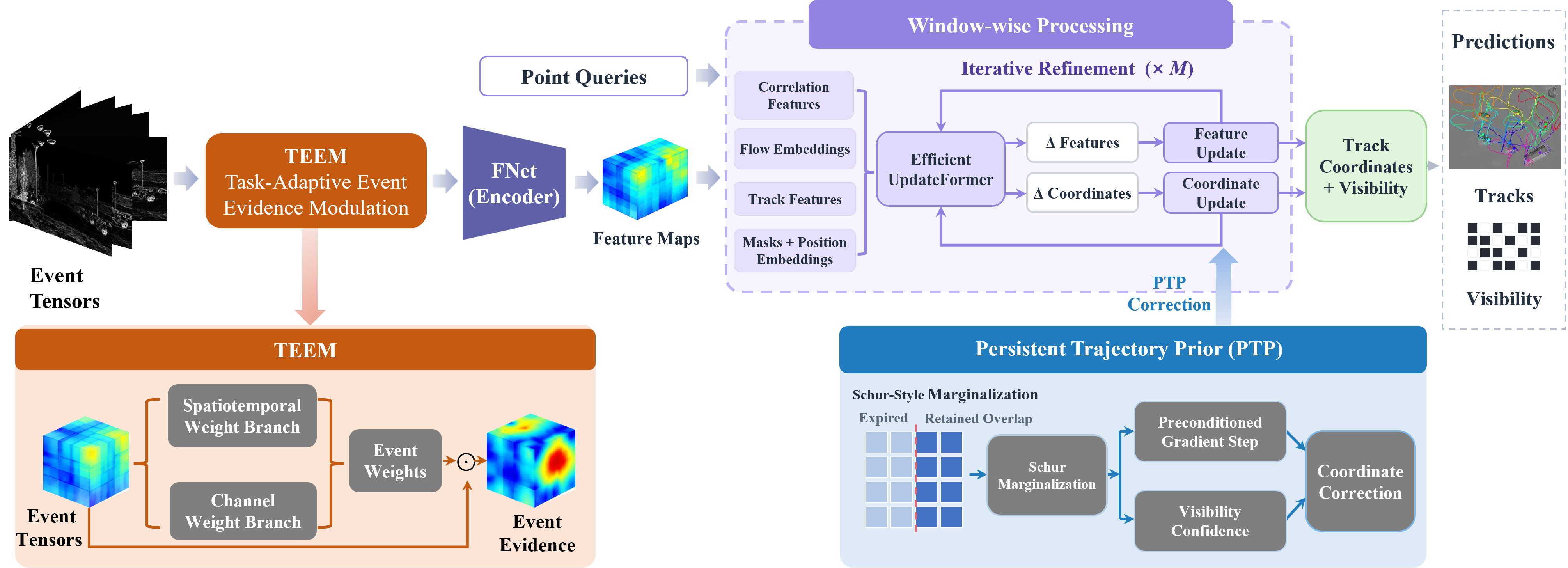}
\caption{Overview of SEPT. TEEM learns spatiotemporal and channel-wise weights for the event tensor before feature extraction. The tracker iteratively refines query trajectories using local correlations and joint attention. At a window transition, PTP transfers overlap coordinates, visibility logits, and a clipped terminal displacement; its fixed Schur precision couples overlap-coordinate deviations during refinement in the next window. Refined latent descriptors are not reused across windows.}
\label{fig:overview}
\end{figure*}

\subsection{Long-Term Feature and Trajectory Tracking}
Long-term tracking must preserve identity when a point moves far from its query location, becomes briefly unobservable, or crosses successive processing windows. Classical feature trackers typically propagate a local estimate and can lose a track once the appearance model or search region is no longer valid. Modern point trackers alleviate this failure in different ways. PIPs and TAPIR repeatedly refine trajectories using temporal context; PIPs++ uses multiple appearance templates and temporally flexible refinement; and CoTracker couples tracks through attention and trains its causal model using windows unrolled over longer sequences \cite{harley2022pips,zheng2023pipsplusplus,doersch2023tapir,karaev2024cotracker,karaev2024cotracker3}. In windowed trackers, the state passed to the next window is often limited to a coordinate initialization or a learned latent representation with implicit geometric meaning.

Fixed-lag estimators provide a complementary principle: when old states leave a window, their information can be marginalized into a prior on the retained states rather than discarded \cite{triggs2000bundle,mourikis2007msckf,leutenegger2015okvis,qin2018vinsmono}. PTP adopts this idea at the trajectory level. It retains overlap coordinates, visibility logits, and a clipped terminal velocity. It uses a small Schur-style precision matrix obtained by marginalizing a fixed local quadratic trajectory model. The resulting prior corrects the retained coordinates throughout the next window's refinement. Compared with high-dimensional feature memory, this design gives the transferred state an explicit geometric meaning, keeps its size bounded per track, and avoids an unbounded gradient path. It is intentionally not a replacement for learned correspondence: new event evidence remains responsible for data association, while PTP supplies continuity when the finite window would otherwise forget solved geometric constraints.

\section{Method}
\label{sec:method}
We first formulate event-based TAP over overlapping windows, then describe the SEPT architecture and its two components.

\subsection{Problem Formulation}
\label{sec:formulation}
An event camera produces an asynchronous stream $\mathcal{E}=\{e_k\}$, where $e_k=(x_k,y_k,\tau_k,p_k)$ records a brightness change at pixel $(x_k,y_k)$ with timestamp $\tau_k$ and polarity $p_k\in\{-1,+1\}$. We evaluate trajectories at discrete timestamps $\{\tau_t\}_{t=1}^{T}$. Following ETAP~\cite{hamann2025etap}, events preceding each timestamp are converted into a mixed-density stack with $C$ channels. Each channel aggregates a nested subset of recent events, providing complementary temporal support. After channel-wise normalization over nonzero entries, the input is denoted by $\vect{V}\in\mathbb{R}^{T\times C\times H\times W}$; the batch dimension is omitted throughout this section.

Given $N$ queries $\mathcal{Q}=\{(t_i,\vect{q}_i)\}_{i=1}^{N}$, with $\vect{q}_i\in\mathbb{R}^{2}$, the tracker predicts coordinates $\hat{\vect{x}}_{t,i}$ and visibility probabilities $\hat v_{t,i}$ for $t\geq t_i$. A query identifies the same physical surface point throughout the sequence, including times when that point is occluded. Processing uses windows $\mathcal{W}_w=\{a_w,\ldots,a_w+S-1\}$, where $a_w=1+ws$, $S$ is the window length, and $s<S$ is the window stride. The overlap between consecutive windows therefore has length $K=S-s$. In our experiments, $C=10$, $S=8$, and $s=4$, giving $K=4$; the remaining numerical settings are reported in Sec.~\ref{sec:implementation}.

\subsection{Overview of SEPT}
\label{sec:overview}
Given the event stack, queries, and overlapping windows defined above, Fig.~\ref{fig:overview} shows how SEPT processes each window and transfers information to the next. TEEM modulates the event stack before the ETAP encoder constructs a feature pyramid. Query-conditioned correlations and a joint space--time transformer then iteratively refine coordinates and descriptors within each window. At a window transition, PTP initializes the next window from the completed trajectory and corrects the overlapping coordinates at each refinement iteration.

\subsubsection{Query-conditioned correspondence}
Let $\widetilde{\vect{V}}$ denote the modulated input. An encoder $\phi$ and successive average pooling produce features $\vect{F}^{l}_t$ at pyramid levels $l=0,\ldots,L-1$. If the encoder stride is $d$, feature-space coordinates are $\vect{z}_{t,i}=\vect{x}_{t,i}/d$. The initial descriptor for query $i$ is bilinearly sampled from $\vect{F}^{0}_{t_i}$ at $\vect{q}_i/d$ and replicated across the active window. At refinement iteration $m$, its current descriptor $\vect{f}_{t,i}^{(m)}\in\mathbb{R}^{D}$ is correlated with features around the current coordinate:
\begin{equation}
c_{t,i}^{l,(m)}(\vect{\delta})
=
\frac{
\left\langle \vect{f}_{t,i}^{(m)},
\vect{F}^{l}_t\!\left(\vect{z}_{t,i}^{(m)}/2^l+\vect{\delta}\right)
\right\rangle}{\sqrt{D}},
\label{eq:correlation}
\end{equation}
where $\vect{\delta}\in\{-r,\ldots,r\}^{2}$ and noninteger locations are bilinearly sampled. Concatenating these values over offsets and levels gives the correlation descriptor $\vect{c}_{t,i}^{(m)}$.

\subsubsection{Iterative trajectory refinement}
The update transformer receives correlation descriptors, displacement embeddings relative to the first time in the window, current point descriptors, query-support indicators, and the window's initial visibility-logit state. Spatial and temporal positional encodings are added to these tokens. Joint attention across trajectories and timestamps produces coordinate and descriptor increments:
\begin{equation}
\begin{aligned}
(\Delta\vect{z}^{(m)},\Delta\vect{f}^{(m)})
&=\Phi_{\theta}(\vect{h}^{(m)}),\\
\vect{z}^{(m+1/2)}
&=\vect{z}^{(m)}+\Delta\vect{z}^{(m)},\\
\vect{f}^{(m+1)}
&=\vect{f}^{(m)}+\mathcal{U}(\Delta\vect{f}^{(m)}),
\end{aligned}
\label{eq:refinement}
\end{equation}
where $\vect{h}^{(m)}$ denotes the assembled tokens and $\mathcal{U}$ is the descriptor-update normalization and multilayer perceptron (MLP). PTP converts the intermediate coordinates $\vect{z}^{(m+1/2)}$ into $\vect{z}^{(m+1)}$ as derived in Sec.~\ref{sec:marg}. After $M$ iterations, coordinates are rescaled to image pixels, and a linear prediction head maps the final descriptors to visibility logits $\ell_{t,i}$, with $\hat v_{t,i}=\sigma(\ell_{t,i})$. Queries that have not been introduced by the end of the active window are excluded from joint attention.

\subsection{Task-Adaptive Event Evidence Modulation}
\label{sec:teem}
Event activity depends on local contrast, motion, and accumulation time. Consequently, a strong event response need not provide a reliable correspondence, and the relative utility of representation channels can change over time. TEEM learns a task-dependent modulation before feature encoding, using complementary local event statistics and global channel statistics.

\subsubsection{Spatiotemporal and channel statistics}
For each input time, we construct three channel-reduced maps:
\begin{equation}
\begin{aligned}
S^{\mathrm{sgn}}_t(x,y)&=\frac{1}{C}\sum_{c=1}^{C}V_{tcxy},\\
S^{\mathrm{mag}}_t(x,y)&=\frac{1}{C}\sum_{c=1}^{C}|V_{tcxy}|,\\
S^{\mathrm{dif}}_t(x,y)&=\frac{1}{C}\sum_{c=1}^{C}
|V_{tcxy}-V_{t-1,cxy}|.
\end{aligned}
\label{eq:event_statistics}
\end{equation}
The difference map is zero at the first input time. The signed and magnitude maps describe complementary aspects of event activity, while the difference map captures changes between adjacent stacks. Their concatenation $\vect{S}\in\mathbb{R}^{T\times3\times H\times W}$ is spatially average-pooled before entering a small 3-D convolutional network:
\begin{equation}
\vect{a}
=
\mathcal{I}\!\left(\psi_{\mathrm{st}}(\mathcal{P}(\vect{S}))\right).
\label{eq:spatial_modulation}
\end{equation}
Here, $\mathcal{P}$ pools spatial dimensions; $\psi_{\mathrm{st}}$ comprises a $3\times3\times3$ convolution, a Gaussian error linear unit (GELU) activation, and a $1\times1\times1$ projection; and $\mathcal{I}$ interpolates to the input resolution. The output $a_{txy}$ is shared by all channels.

In parallel, we compute per-channel spatial means,
\begin{equation}
g_{tc}^{\mathrm{sgn}}=\frac{1}{HW}\sum_{x,y}V_{tcxy},
\qquad
g_{tc}^{\mathrm{mag}}=\frac{1}{HW}\sum_{x,y}|V_{tcxy}|.
\label{eq:channel_statistics}
\end{equation}
A two-layer MLP maps the concatenated $2C$ statistics at each time to channel logits $\vect{b}_t\in\mathbb{R}^{C}$. Thus, the spatiotemporal branch models local activity and its temporal variation, while the channel branch adjusts the balance among the stack's temporal supports.

\subsubsection{Residual modulation and scale normalization}
The two branches are combined through a bounded residual gate,
\begin{equation}
\widehat W_{tcxy}
=
1+\alpha\tanh(a_{txy}+b_{tc}),
\qquad 0\leq\alpha\leq1.
\label{eq:weight}
\end{equation}
The unnormalized gate satisfies $1-\alpha\leq\widehat W\leq1+\alpha$. To stabilize the aggregate input scale, we normalize it over active entries. Let $\Omega_t=\{(c,x,y):V_{tcxy}\neq0\}$ and define
\begin{equation}
\nu_t=
\begin{cases}
|\Omega_t|^{-1}\displaystyle\sum_{(c,x,y)\in\Omega_t}
\widehat W_{tcxy}, & |\Omega_t|>0,\\[1mm]
(C H W)^{-1}\displaystyle\sum_{c,x,y}\widehat W_{tcxy},
& |\Omega_t|=0.
\end{cases}
\label{eq:weight_scale}
\end{equation}
The modulated representation is
\begin{equation}
W_{tcxy}=\frac{\widehat W_{tcxy}}{\max(\nu_t,\varepsilon_{\mathrm{w}})},
\qquad
\widetilde V_{tcxy}=W_{tcxy}V_{tcxy}.
\label{eq:normalized_modulation}
\end{equation}
The bound in Eq.~\eqref{eq:weight} applies before normalization; Eq.~\eqref{eq:normalized_modulation} retains relative modulation while stabilizing the mean gain on active responses. Multiplication also preserves zero entries in the input. Both output projections are initialized to zero, so that $\widehat W=W=1$ initially. TEEM therefore begins as an identity mapping and learns to redistribute event evidence through downstream tracking supervision, without event-importance annotations.

\subsection{Persistent Trajectory Prior}
\label{sec:marg}
Overlapping windows allow coordinates to be copied into the next window, but copying alone does not define how subsequent updates should respect the previous trajectory. PTP supplies this missing constraint with a compact quadratic prior on overlap deviations. It combines a fixed temporal precision with the completed window's coordinates and predicted visibility.

\subsubsection{Marginalizing a local trajectory model}
Consider one point and one coordinate component in a completed window. Let $\vect{r}\in\mathbb{R}^{S}$ denote a perturbation about its refined trajectory. We define
\begin{equation}
\begin{aligned}
E(\vect{r})
&=\frac{\lambda_o}{2}\|\vect{r}\|_2^2
+\frac{\lambda_s}{2}\|\vect{D}_2\vect{r}\|_2^2\\
&=\frac{1}{2}\vect{r}^{\top}\vect{H}\vect{r},
\qquad
\vect{H}=\lambda_o\vect{I}+\lambda_s\vect{D}_2^{\top}\vect{D}_2,
\end{aligned}
\label{eq:trajectory_energy}
\end{equation}
where $(\vect{D}_2\vect{r})_j=r_j-2r_{j+1}+r_{j+2}$, $\lambda_o>0$, and $\lambda_s\geq0$. The first term anchors the completed solution; the second couples changes to that solution across time. Because the model acts on perturbations, it does not require the reference trajectory itself to have constant velocity.

\begin{figure*}[t]
\centering
\includegraphics[width=0.9\textwidth]{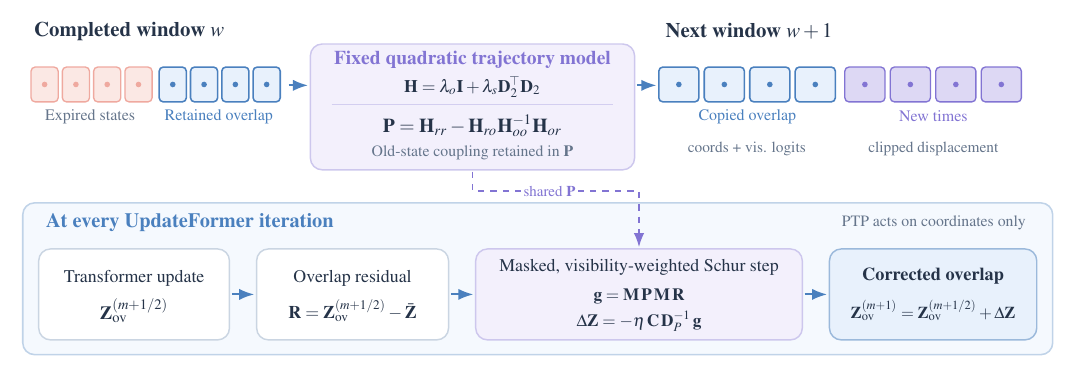}
\caption{Schur-style persistent trajectory prior. Expired states are eliminated from a fixed quadratic perturbation model, leaving precision $\vect{P}$ over the retained overlap. The next window copies overlap coordinates and visibility logits and initializes coordinates at newly exposed timestamps using a clipped terminal displacement. After each transformer update, a validity-masked and visibility-weighted preconditioned gradient softly corrects only overlap coordinates; coordinates at newly exposed timestamps receive no prior correction. State counts are illustrative.}
\label{fig:schur_visualization}
\end{figure*}

Partition $\vect{r}=[\vect{r}_o^{\top},\vect{r}_r^{\top}]^{\top}$ into the first $s$ states, which leave the window, and the last $K$ states, which remain. Minimization over the expired states gives
\begin{equation}
\vect{r}_o^{*}
=
-\vect{H}_{oo}^{-1}\vect{H}_{or}\vect{r}_r.
\label{eq:expired_elimination}
\end{equation}
Substituting this solution into Eq.~\eqref{eq:trajectory_energy} yields the reduced energy
\begin{equation}
\begin{aligned}
E_{\mathrm{red}}(\vect{r}_r)
&=\frac{1}{2}\vect{r}_r^{\top}\vect{P}\vect{r}_r,\\
\vect{P}
&=\vect{H}_{rr}
-\vect{H}_{ro}\vect{H}_{oo}^{-1}\vect{H}_{or}.
\end{aligned}
\label{eq:schur}
\end{equation}
Equivalently, $\vect{P}$ is the precision obtained by marginalizing the zero-mean Gaussian associated with Eq.~\eqref{eq:trajectory_energy}. This reduction preserves the temporal coupling specified by the fixed quadratic model. The resulting precision is shared across points and is independent of visual residuals; the scene-dependent reference comes from the completed trajectory. Since the model and window geometry are fixed, $\vect{P}$ and its diagonal preconditioner are computed once. Fig.~\ref{fig:schur_visualization} summarizes the state partition, Schur elimination, and overlap-only correction applied during iterative refinement.

\subsubsection{State transfer and window initialization}
Let $\hat{\vect{z}}_{j,i}$ and $\hat\ell_{j,i}$ denote the completed window's feature-space coordinates and visibility logits at local time $j$. PTP retains
\begin{equation}
\bar{\vect{z}}_{j,i}=\hat{\vect{z}}_{s+j,i},
\qquad
\bar\ell_{j,i}=\hat\ell_{s+j,i},
\quad j=1,\ldots,K.
\label{eq:overlap_state}
\end{equation}
It also computes a clipped terminal displacement
\begin{equation}
\vect{u}_i
=
\operatorname{clip}_{u_{\max}}
\!\left(\hat{\vect{z}}_{S,i}-\hat{\vect{z}}_{S-1,i}\right),
\label{eq:terminal_velocity}
\end{equation}
where $\operatorname{clip}_{u_{\max}}$ limits the vector's Euclidean norm to $u_{\max}$. For a query carried from the preceding window, the next coordinate initialization is
\begin{equation}
\vect{z}_{j,i}^{(0)}
=
\begin{cases}
\bar{\vect{z}}_{j,i}, & 1\leq j\leq K,\\
\bar{\vect{z}}_{K,i}+(j-K)\vect{u}_i, & K<j\leq S.
\end{cases}
\label{eq:prior_initialization}
\end{equation}
Visibility logits are copied to the overlap, and the terminal logit is repeated at newly exposed timestamps. New queries retain the query-based initialization. Stored coordinates, logits, and terminal displacements are detached at the boundary, and refined latent descriptors are not propagated into the next window. This keeps the transferred state bounded per track while retaining a geometric reference for the next refinement.

\subsubsection{Visibility-aware overlap correction}
PTP also constrains the overlap after every transformer update. For point $i$, collect the current intermediate overlap coordinates into $\vect{Z}_i^{(m+1/2)}\in\mathbb{R}^{K\times2}$ and the stored reference into $\bar{\vect{Z}}_i$. Let $\vect{M}_i$ be a diagonal validity mask: an entry is one only when its timestamp lies in the sequence and is no earlier than the query time. The residual and its masked prior energy are
\begin{equation}
\begin{aligned}
\vect{R}_i^{(m)}
&=\vect{Z}_i^{(m+1/2)}-\bar{\vect{Z}}_i,\\
E_i^{(m)}
&=\frac{1}{2}\operatorname{tr}\!\left(
(\vect{R}_i^{(m)})^{\top}
\vect{M}_i\vect{P}\vect{M}_i\vect{R}_i^{(m)}
\right).
\end{aligned}
\label{eq:masked_prior_energy}
\end{equation}
The gradient is $\vect{M}_i\vect{P}\vect{M}_i\vect{R}_i^{(m)}$. We precondition this gradient by the diagonal of $\vect{P}$ and scale it using the previous window's visibility:
\begin{equation}
\begin{aligned}
\vect{D}_P&=\operatorname{diag}\!\left(
\{\max(P_{jj},\varepsilon_{\mathrm{p}})\}_{j=1}^{K}\right),\\
\vect{C}_i&=\operatorname{diag}\!\left(
\{c_{\min}+(1-c_{\min})\sigma(\bar\ell_{j,i})\}_{j=1}^{K}
\right).
\end{aligned}
\label{eq:prior_preconditioning}
\end{equation}
The resulting coordinate correction is
\begin{equation}
\Delta\vect{Z}_{i,\mathrm{prior}}^{(m)}
=
-\eta\,\vect{C}_i\vect{D}_P^{-1}
\vect{M}_i\vect{P}\vect{M}_i\vect{R}_i^{(m)},
\label{eq:prior}
\end{equation}
where $\eta$ controls the update magnitude. The overlap update is
\begin{equation}
\vect{Z}_i^{(m+1)}
=
\vect{Z}_i^{(m+1/2)}
+
\Delta\vect{Z}_{i,\mathrm{prior}}^{(m)}.
\label{eq:overlap_update}
\end{equation}
Coordinates outside the overlap receive only the transformer update. When no preceding window is available, the prior correction is zero.

The visibility factor weakens the constraint on previously uncertain tracks, while $c_{\min}>0$ retains a minimum influence. Retaining logits for initialization and applying the sigmoid only for confidence weighting keeps the transferred state in consistent units. The prior is soft: new event evidence can move coordinates away from the stored trajectory. PTP introduces no trainable parameters, and its per-window correction costs $O(NK^2)$ with a shared $K\times K$ precision matrix.

\subsection{Training Objective}
\label{sec:training_objective}
SEPT is supervised by ground-truth coordinates $\vect{x}^{*}_{t,i}$ and visibility labels $y_{t,i}\in\{0,1\}$. Let $\mathcal{A}_w$ contain the valid point--time pairs in window $w$ that are no earlier than the corresponding query, and let $n_w=\max(|\mathcal{A}_w|,1)$. Occluded coordinates receive weight $\omega_{t,i}=y_{t,i}+\beta(1-y_{t,i})$, with $\beta=0.2$. For $N_w$ training windows and $M$ refinements, the coordinate loss is
\begin{equation}
\begin{aligned}
\mathcal{L}_{\mathrm{track}}
={}&\frac{1}{N_wM}\sum_{w=1}^{N_w}\sum_{m=1}^{M}
\frac{\gamma^{M-m}}{n_w}\\
&\quad\cdot\sum_{(t,i)\in\mathcal{A}_w}
\omega_{t,i}
\frac{\|\hat{\vect{x}}^{w,(m)}_{t,i}-\vect{x}^{*}_{t,i}\|_1}{2},
\end{aligned}
\label{eq:track_loss}
\end{equation}
where $\gamma=0.8$ gives greater weight to later refinements. The factor $1/2$ averages the two coordinate components. Each window's final visibility prediction is supervised by binary cross-entropy (BCE):
\begin{equation}
\mathcal{L}_{\mathrm{vis}}
=
\frac{1}{N_w}\sum_{w=1}^{N_w}\frac{1}{n_w}
\sum_{(t,i)\in\mathcal{A}_w}
\operatorname{BCE}(\hat v^w_{t,i},y_{t,i}).
\label{eq:visibility_loss}
\end{equation}
Probabilities are clipped away from zero and one for numerical stability. The total objective is
\begin{equation}
\mathcal{L}=0.1\mathcal{L}_{\mathrm{track}}+\mathcal{L}_{\mathrm{vis}}.
\label{eq:total_loss}
\end{equation}
Supervision is applied to each window, including its overlap. TEEM receives gradients through event encoding and tracking, while PTP modifies the forward coordinate refinement without an additional prior loss. Coordinates are detached between refinement iterations, and carried states are detached between windows. Training uses the same coordinate and visibility supervision as our retrained baseline, without time-inverted event pairs or contrastive feature-alignment supervision~\cite{hamann2025etap}.

\section{Experiments}
\label{sec:experiments}
We evaluate SEPT on two complementary tasks: TAP and feature tracking. After describing the implementation and synthetic training setup, we report results for both tasks and analyze the contributions of TEEM and PTP.

\subsection{Implementation Details}
\label{sec:implementation}
We train on three NVIDIA GeForce RTX 4090 GPUs, each with 24 GB of memory, using distributed data parallelism. The batch size is one sample per GPU, giving a global batch size of three. Each training sample contains $T=24$ event tensors with $C=10$ channels at $H=W=512$ resolution and $N=256$ randomly sampled trajectories. The encoder stride is $d=4$, the feature dimension is $D=128$, and query correlation uses $L=4$ pyramid levels with radius $r=3$. We use windows of length $S=8$ with stride $s=4$ (overlap length $K=4$) and perform $M=4$ refinement iterations.

We use bfloat16 mixed precision and AdamW with a learning rate of $5\times10^{-4}$, a minimum learning rate of $5\times10^{-5}$, weight decay of $10^{-4}$, and 170\,000 optimization steps. A cosine schedule decays the learning rate to its minimum over training. 

For the two proposed modules, TEEM uses a hidden dimension of eight, a spatial downsampling factor of four, modulation strength $\alpha=1$, and normalization floor $\varepsilon_{\mathrm{w}}=10^{-4}$. PTP uses $\lambda_o=\lambda_s=1$, update weight $\eta=0.25$, confidence floor $c_{\min}=0.10$, precision-diagonal floor $\varepsilon_{\mathrm{p}}=10^{-6}$, and terminal-displacement clip $u_{\max}=4$ feature-map pixels. No auxiliary TEEM or PTP regularization is applied; all variants share the same coordinate and visibility losses.

\subsection{Synthetic Training Data and Baseline}
\label{sec:training_data}
\begin{figure}[t]
\centering
\includegraphics[width=\columnwidth]{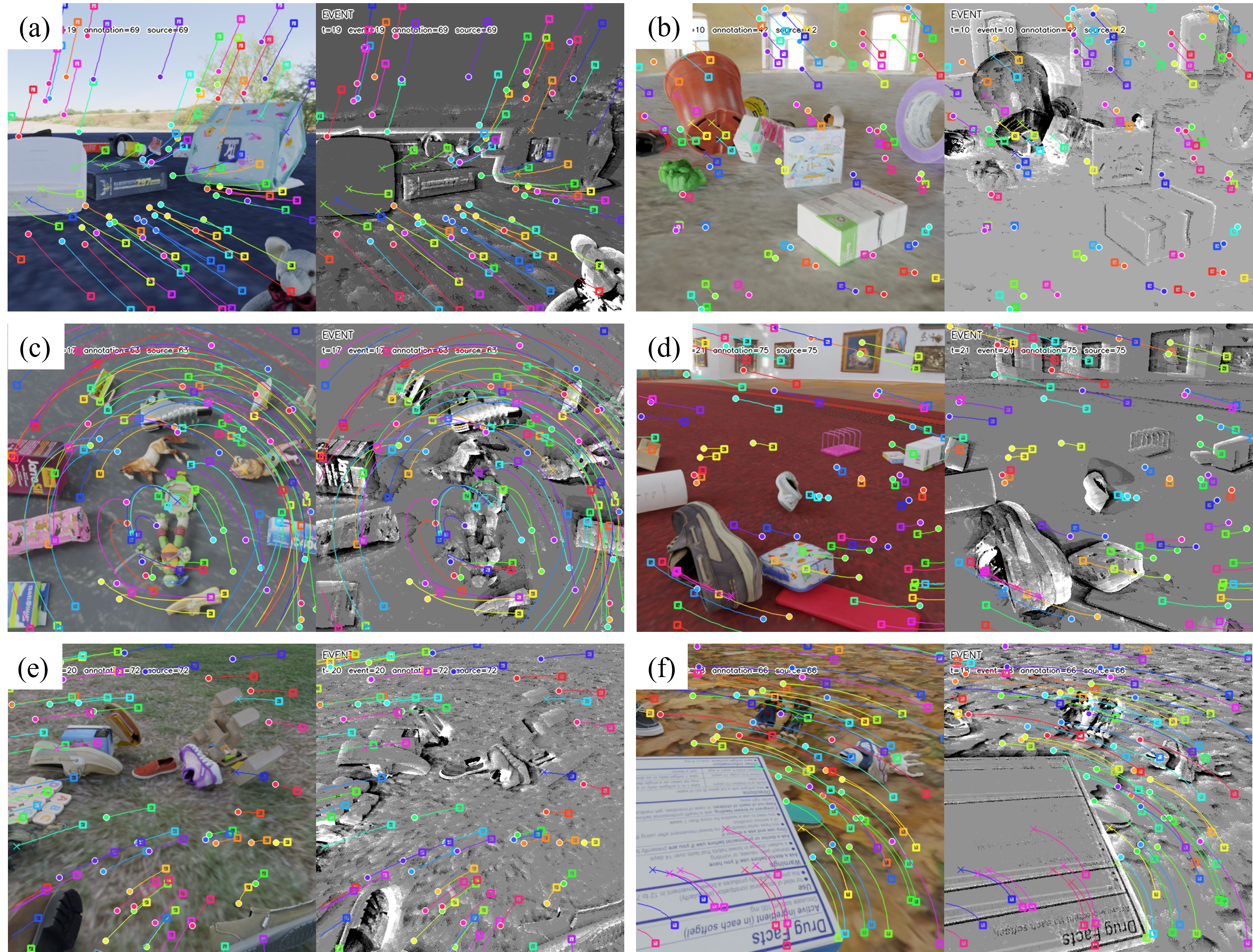}
\caption{Synthetic EventKubric examples with varied motion. In each pair (a)--(f), the left panel shows the rendered RGB frame and the right panel its event representation. Colored markers and curves indicate query locations and ground-truth trajectories; an $\times$ marks a query point that is occluded at the displayed time.}
\label{fig:synthetic_data}
\end{figure}

We use 7021 synthetic EventKubric samples generated following the ETAP pipeline~\cite{hamann2025etap}. Dynamic scenes are constructed with Kubric~\cite{greff2022kubric} and rendered as red-green-blue (RGB) video using Blender. Following the VID2E video-to-events workflow~\cite{gehrig2020vid2e}, the rendered frames are temporally interpolated and converted into synthetic event streams. Point trajectories and visibility labels are derived from Kubric's depth, segmentation, and surface-normal annotations. Fig.~\ref{fig:synthetic_data} shows six examples with varied motion; the RGB frames visualize the scenes, while the event representations are the tracker inputs. The default training/validation/test allocation is $80{:}15{:}5$; dagger-marked variants use a $90{:}5{:}5$ allocation of the same corpus. Thus, the dagger indicates a larger training subset rather than newly generated scenes.

All variants are trained exclusively on synthetic event data. ETAP denotes the baseline retrained from its released source code under our training protocol. We omit time-inverted event pairs and the associated contrastive feature-alignment loss for both the baseline and SEPT. This baseline comparison isolates the architectural modifications under a common event-only training protocol.

\subsection{Task 1: Tracking Any Point}
\label{sec:tap_results}
\subsubsection{Datasets and metrics}
We evaluate TAP on the synthetic EventKubric test set~\cite{hamann2025etap} and EVIMO2~\cite{burner2022evimo2}, an event-camera benchmark for motion segmentation, optical flow, structure from motion, and visual-inertial odometry in indoor scenes. EventKubric assesses correspondence and visibility estimation in simulated scenes, while EVIMO2 tests transfer to real event observations. We report three metrics introduced by TAP-Vid~\cite{doersch2022tap}: average Jaccard (AJ), average points within threshold ($\delta^x_{\mathrm{avg}}$), and occlusion accuracy (OA). AJ evaluates localization jointly with visibility, $\delta^x_{\mathrm{avg}}$ measures visible-point localization, and OA measures visibility classification. All scores are on $[0,1]$, with higher values indicating better performance.

\begin{table}[t]
\centering
\footnotesize
\caption{TAP results on EventKubric and EVIMO2. ETAP~\cite{hamann2025etap} is our retrained baseline; SEPT combines TEEM and PTP.}
\label{tab:tap_main}
\setlength{\tabcolsep}{3pt}
\begin{tabular}{llccc}
\toprule
Dataset & Method & AJ $\uparrow$ & $\delta^x_{\mathrm{avg}}$ $\uparrow$ & OA $\uparrow$\\
\midrule
EventKubric & ETAP$^\dagger$ & 0.512 & 0.633 & 0.951\\
 & SEPT$^\dagger$ & \textbf{0.578} & \textbf{0.693} & \textbf{0.955}\\
\midrule
EVIMO2 & ETAP$^\dagger$ & 0.635 & 0.751 & \textbf{0.943}\\
 & SEPT$^\dagger$ & \textbf{0.673} & \textbf{0.786} & 0.936\\
\bottomrule
\end{tabular}
\end{table}

\subsubsection{Results on EventKubric}
On the synthetic EventKubric test set, SEPT raises AJ from $0.512$ to $0.578$ and the visible-point localization score $\delta^x_{\mathrm{avg}}$ from $0.633$ to $0.693$ (Table~\ref{tab:tap_main}). OA is already high for ETAP and changes only from $0.951$ to $0.955$. The larger localization gain indicates that SEPT primarily improves point correspondence over time while preserving strong visibility prediction.

\subsubsection{Results on EVIMO2}
On the real EVIMO2 sequences, SEPT raises AJ from $0.635$ to $0.673$ and $\delta^x_{\mathrm{avg}}$ from $0.751$ to $0.786$, although OA decreases from $0.943$ to $0.936$ (Table~\ref{tab:tap_main}). The AJ gain despite this small visibility penalty shows that the localization benefit transfers from synthetic training to real events without fine-tuning. Across both benchmarks, improved localization is the consistent source of SEPT's TAP gain; the small OA decline on EVIMO2 identifies real-data visibility estimation as the remaining weakness.

\subsection{Task 2: Feature Tracking}
\label{sec:feature_results}
\subsubsection{Datasets and metrics}
We evaluate feature tracking on the Event Camera (EC) dataset \cite{mueggler2017event} and the Event-aided Direct Sparse Odometry (EDS) dataset \cite{hidalgocarrio2022eds} using the data and protocol of ETAP~\cite{hamann2025etap}; we additionally evaluate on the Multi-Robot, Multi-Sensor, Multi-Environment Event Dataset (M3ED) \cite{chaney2023m3ed}. Feature age (FA) measures normalized tracking lifetime among tracks with positive lifetime; expected feature age (EFA) also accounts for immediately failed tracks~\cite{messikommer2023ddft}. Following the same evaluation protocol, we average both metrics over localization thresholds from 1 to 31 pixels. Higher FA and EFA are better.

For M3ED, we construct feature-tracking ground truth from the provided depth maps and camera poses. Candidate image points with valid depth are back-projected into 3-D and reprojected through the camera trajectory to obtain their 2-D positions over time. Because this geometric construction assumes that a point is stationary in the world, we manually review the proposed trajectories and exclude points on moving objects or with unreliable projections before accepting them as ground truth. Points that leave the field of view or become occluded retain their identities but are marked invisible and excluded from position-error evaluation. The selected sequences span vehicle-mounted and unmanned aerial vehicle recordings, daytime and nighttime conditions, and urban, indoor, and outdoor scenes, providing a diverse test of transfer beyond the EC and EDS benchmarks. Each evaluated segment lasts 2 s and is sampled at 25 Hz, yielding 51 timestamps. For FA and EFA, each point's evaluable span is its initial uninterrupted ground-truth-visible interval; the first localization error beyond the selected threshold ends its predicted lifetime.

Because the RGB and event sensors in our M3ED setup are not spatially co-registered, we exclude RGB--event fusion methods that require aligned inputs.

\begin{table}[t]
\centering
\scriptsize
\caption{Feature tracking on EC and EDS. E denotes events; E+F denotes events and frames. MATE results are quoted from its original paper. M3ED results are reported by sequence in Table~\ref{tab:m3ed_per_sequence}.}
\label{tab:feature_main}
\setlength{\tabcolsep}{2.3pt}
\begin{tabular}{llcccc}
\toprule
\multirow{2}{*}{Method} & \multirow{2}{*}{Input} & \multicolumn{2}{c}{EC} & \multicolumn{2}{c}{EDS}\\
\cmidrule(lr){3-4}\cmidrule(lr){5-6}
& & FA & EFA & FA & EFA\\
\midrule
ICP~\cite{kueng2016lowlatency} & E & 0.256 & 0.245 & 0.060 & 0.040\\
EKLT~\cite{gehrig2019eklt} & E+F & 0.811 & 0.775 & 0.325 & 0.205\\
DDFT~\cite{messikommer2023ddft} & E+F & 0.825 & 0.818 & 0.576 & 0.472\\
FE-TAP~\cite{liu2024fetap} & E+F & 0.844 & 0.838 & 0.676 & 0.589\\
EM-ICP~\cite{zhu2017probabilistic} & E & 0.337 & 0.334 & 0.161 & 0.120\\
HASTE~\cite{alzugaray2020haste} & E & 0.442 & 0.427 & 0.096 & 0.063\\
DDFT E2VID~\cite{messikommer2023ddft,rebecq2019e2vid} & E & 0.794 & 0.786 & 0.589 & 0.495\\
MATE~\cite{han2025mate} & E & \textbf{0.885} & \textbf{0.875} & 0.713 & \textbf{0.626}\\
\midrule
ETAP$^\dagger$~\cite{hamann2025etap} & E & 0.855 & 0.850 & 0.711 & 0.605\\
SEPT$^\dagger$ (ours) & E & 0.866 & 0.861 & \textbf{0.732} & 0.619\\
\bottomrule
\end{tabular}
\end{table}

\subsubsection{Results on EC and EDS}
Table~\ref{tab:feature_main} compares SEPT with event-only methods and methods combining events and frames. Published baseline scores are reproduced from ETAP~\cite{hamann2025etap}, except for MATE, whose scores are quoted from its original paper~\cite{han2025mate}. MATE's official implementation and pretrained weights were unavailable to us, so we could not re-evaluate it under our pipeline. Since its training conditions differ from ours, its scores provide a reference comparison rather than a controlled same-protocol comparison. The ETAP$^\dagger$ and SEPT$^\dagger$ results are from our own runs. SEPT achieves $0.866$ FA and $0.861$ EFA on EC, exceeding our retrained ETAP scores of $0.855$ and $0.850$. On EDS, SEPT improves FA from $0.711$ to $0.732$ and EFA from $0.605$ to $0.619$. Compared with MATE's published results, SEPT achieves higher FA on EDS, while MATE reports higher FA and EFA on EC and higher EFA on EDS.

\subsubsection{Results on M3ED}
\begin{figure*}[!t]
\centering
\includegraphics[width=\textwidth]{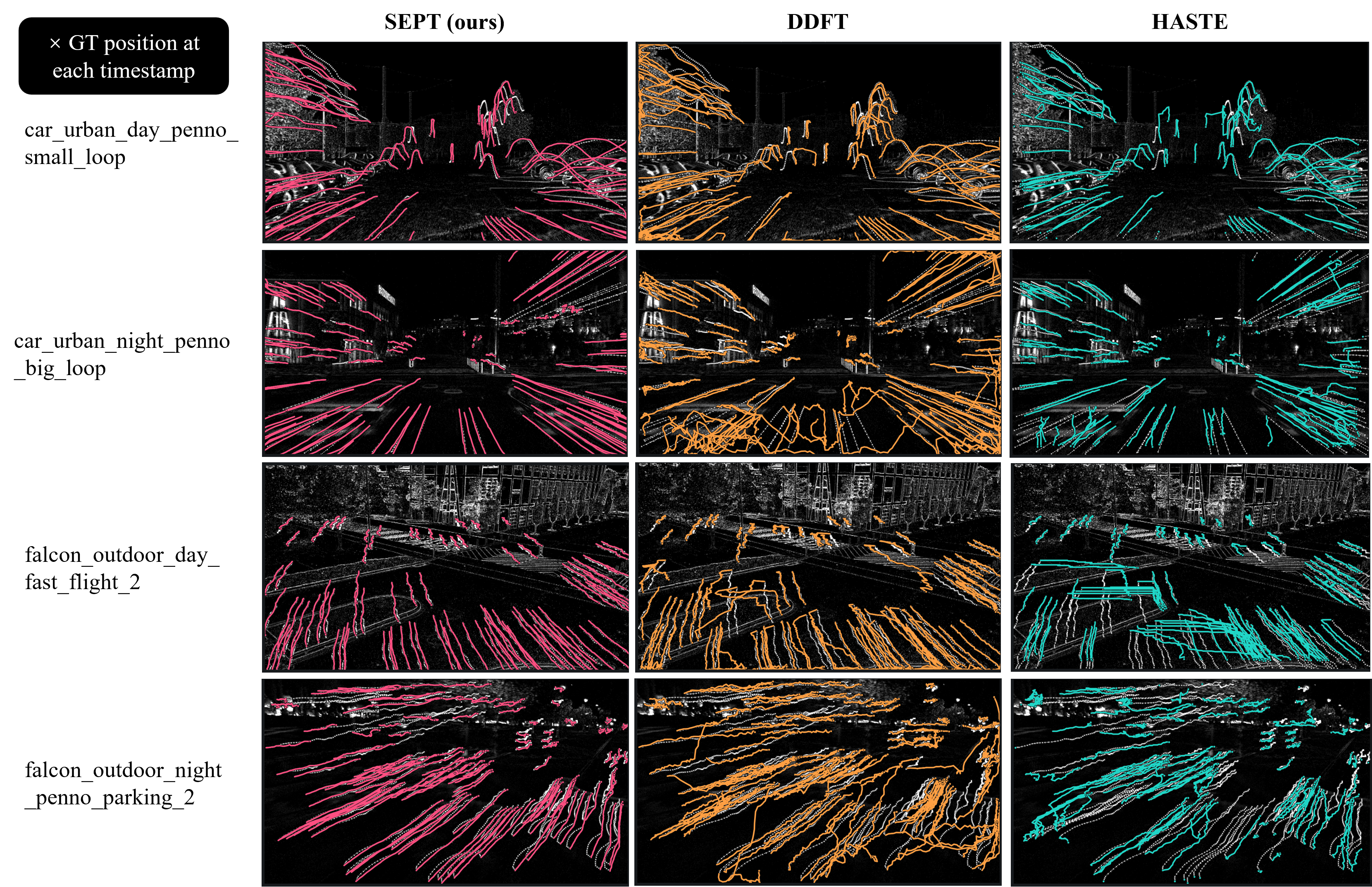}
\caption{Qualitative feature-tracking results on M3ED. Rows show four sequences; columns compare SEPT, DDFT~\cite{messikommer2023ddft}, and HASTE~\cite{alzugaray2020haste} from left to right. Colored curves show predicted tracks, and $\times$ marks the ground-truth point position at each timestamp.}
\label{fig:m3ed_qualitative}
\end{figure*}

Table~\ref{tab:m3ed_per_sequence} reports FA and EFA for each of the nine evaluated M3ED sequences. Averaged over sequences, SEPT reaches $0.797$ FA and $0.795$ EFA, compared with $0.795$ and $0.792$ for the retrained ETAP baseline. SEPT therefore improves both feature-age measures on this real benchmark despite training exclusively on synthetic EventKubric data. We attribute the smaller margin to differences in event content and motion between the synthetic training scenes and M3ED, which make the learned TEEM weighting harder to transfer. Fig.~\ref{fig:m3ed_qualitative} qualitatively compares trajectories from SEPT, DDFT, and HASTE on four M3ED sequences.

\begin{table*}[!t]
\centering
\scriptsize
\caption{Per-sequence feature tracking on the nine evaluated M3ED left-camera segments. Entries are FA/EFA, averaged over localization thresholds from 1 to 31 pixels. DDFT~\cite{messikommer2023ddft} uses its event-only configuration. The final row is the unweighted mean across sequences.}
\label{tab:m3ed_per_sequence}
\setlength{\tabcolsep}{4pt}
\renewcommand{\arraystretch}{1.06}
\begin{tabular*}{\textwidth}{@{\extracolsep{\fill}}lccccc@{}}
\toprule
Sequence & HASTE~\cite{alzugaray2020haste} & AEB-Tracker~\cite{wang2024aeb} & DDFT~\cite{messikommer2023ddft} & ETAP$^\dagger$ & SEPT$^\dagger$ (ours)\\
\midrule
\shortstack[l]{\texttt{car\_urban\_day\_}\\\texttt{penno\_small\_loop}} & 0.733/0.669 & 0.530/0.455 & 0.813/0.807 & \textbf{0.831}/\textbf{0.829} & 0.824/0.822\\
\shortstack[l]{\texttt{car\_urban\_night\_}\\\texttt{penno\_big\_loop}} & 0.693/0.632 & 0.553/0.504 & 0.683/0.657 & 0.836/0.828 & \textbf{0.854}/\textbf{0.845}\\
\shortstack[l]{\texttt{car\_urban\_night\_}\\\texttt{penno\_small\_loop}} & 0.577/0.541 & 0.457/0.429 & 0.557/0.540 & \textbf{0.736}/\textbf{0.735} & 0.731/0.729\\
\shortstack[l]{\texttt{falcon\_indoor\_}\\\texttt{flight\_1}} & 0.576/0.502 & 0.426/0.387 & 0.652/0.644 & 0.740/0.739 & \textbf{0.742}/\textbf{0.741}\\
\shortstack[l]{\texttt{falcon\_indoor\_}\\\texttt{flight\_2}} & 0.611/0.558 & 0.539/0.482 & 0.750/0.741 & \textbf{0.827}/\textbf{0.825} & 0.802/0.801\\
\shortstack[l]{\texttt{falcon\_outdoor\_day\_}\\\texttt{fast\_flight\_2}} & 0.614/0.606 & 0.434/0.396 & 0.799/0.704 & 0.878/0.877 & \textbf{0.881}/\textbf{0.881}\\
\shortstack[l]{\texttt{falcon\_outdoor\_night\_}\\\texttt{high\_beams}} & 0.673/0.664 & 0.353/0.349 & 0.380/0.376 & \textbf{0.860}/\textbf{0.859} & 0.854/0.853\\
\shortstack[l]{\texttt{falcon\_outdoor\_night\_}\\\texttt{penno\_parking\_1}} & 0.428/0.383 & 0.361/0.329 & 0.452/0.441 & 0.747/0.742 & \textbf{0.777}/\textbf{0.773}\\
\shortstack[l]{\texttt{falcon\_outdoor\_night\_}\\\texttt{penno\_parking\_2}} & 0.501/0.452 & 0.322/0.316 & 0.493/0.489 & 0.697/0.695 & \textbf{0.710}/\textbf{0.709}\\
\midrule
Mean & 0.601/0.556 & 0.441/0.405 & 0.620/0.600 & 0.795/0.792 & \textbf{0.797}/\textbf{0.795}\\
\bottomrule
\end{tabular*}
\end{table*}

\subsection{Ablation and Training-Data Sensitivity}

\label{sec:ablations}
We first analyze TEEM and PTP individually, then their combined performance and inference cost, and finally their sensitivity to the synthetic training-data allocation. Table~\ref{tab:ablation_all} reports ETAP, the two single-module variants, and SEPT on all five datasets under both training allocations.

\begin{table*}[!t]
\centering
\footnotesize
\caption{Module ablation and training-data sensitivity across TAP (EventKubric and EVIMO2) and feature tracking (EC, EDS, and M3ED). Bold marks the best score for each dataset and metric within a training split.}
\label{tab:ablation_all}
\setlength{\tabcolsep}{2.5pt}
\begin{tabular*}{\textwidth}{@{\extracolsep{\fill}}l*{12}{c}@{}}
\toprule
 & \multicolumn{6}{c}{TAP} & \multicolumn{6}{c}{Feature tracking}\\
\cmidrule(lr){2-7}\cmidrule(lr){8-13}
 & \multicolumn{3}{c}{EventKubric} & \multicolumn{3}{c}{EVIMO2} & \multicolumn{2}{c}{EC} & \multicolumn{2}{c}{EDS} & \multicolumn{2}{c}{M3ED}\\
\cmidrule(lr){2-4}\cmidrule(lr){5-7}\cmidrule(lr){8-9}\cmidrule(lr){10-11}\cmidrule(lr){12-13}
Model & AJ & $\delta^x_{\mathrm{avg}}$ & OA & AJ & $\delta^x_{\mathrm{avg}}$ & OA & FA & EFA & FA & EFA & FA & EFA\\
\midrule
\multicolumn{13}{l}{\textbf{Standard training split (80:15:5)}}\\
\midrule
ETAP & 0.524 & 0.643 & \textbf{0.949} & 0.622 & 0.737 & 0.940 & 0.839 & 0.833 & 0.699 & 0.599 & 0.789 & 0.787\\
ETAP+TEEM & 0.531 & 0.649 & 0.944 & 0.655 & 0.771 & 0.937 & 0.797 & 0.792 & 0.719 & 0.605 & 0.795 & \textbf{0.795}\\
ETAP+PTP & \textbf{0.566} & \textbf{0.681} & 0.946 & \textbf{0.663} & \textbf{0.777} & \textbf{0.943} & \textbf{0.866} & \textbf{0.861} & \textbf{0.726} & 0.614 & \textbf{0.797} & 0.794\\
SEPT & 0.554 & 0.676 & 0.948 & 0.643 & 0.761 & 0.930 & 0.854 & 0.849 & 0.722 & \textbf{0.615} & \textbf{0.797} & 0.794\\
\midrule
\multicolumn{13}{l}{\textbf{Expanded training split (90:5:5)$^\dagger$}}\\
\midrule
ETAP & 0.512 & 0.633 & 0.951 & 0.635 & 0.751 & \textbf{0.943} & 0.855 & 0.850 & 0.711 & 0.605 & 0.795 & 0.792\\
ETAP+TEEM & 0.524 & 0.643 & 0.943 & 0.652 & 0.762 & 0.937 & 0.869 & 0.864 & 0.686 & 0.587 & 0.792 & 0.789\\
ETAP+PTP & 0.534 & 0.652 & 0.939 & 0.645 & 0.760 & 0.935 & \textbf{0.874} & \textbf{0.869} & 0.725 & 0.609 & \textbf{0.797} & 0.794\\
SEPT & \textbf{0.578} & \textbf{0.693} & \textbf{0.955} & \textbf{0.673} & \textbf{0.786} & 0.936 & 0.866 & 0.861 & \textbf{0.732} & \textbf{0.619} & \textbf{0.797} & \textbf{0.795}\\
\bottomrule
\end{tabular*}
\end{table*}

\subsubsection{Contribution of TEEM}
ETAP+TEEM increases AJ over ETAP on both EventKubric and EVIMO2 under each training allocation (Table~\ref{tab:ablation_all}). Its feature-tracking effect varies by dataset, indicating that learned event weighting is more sensitive to the event distribution when used alone.

\begin{figure*}[!t]
\centering
\includegraphics[width=0.97\textwidth]{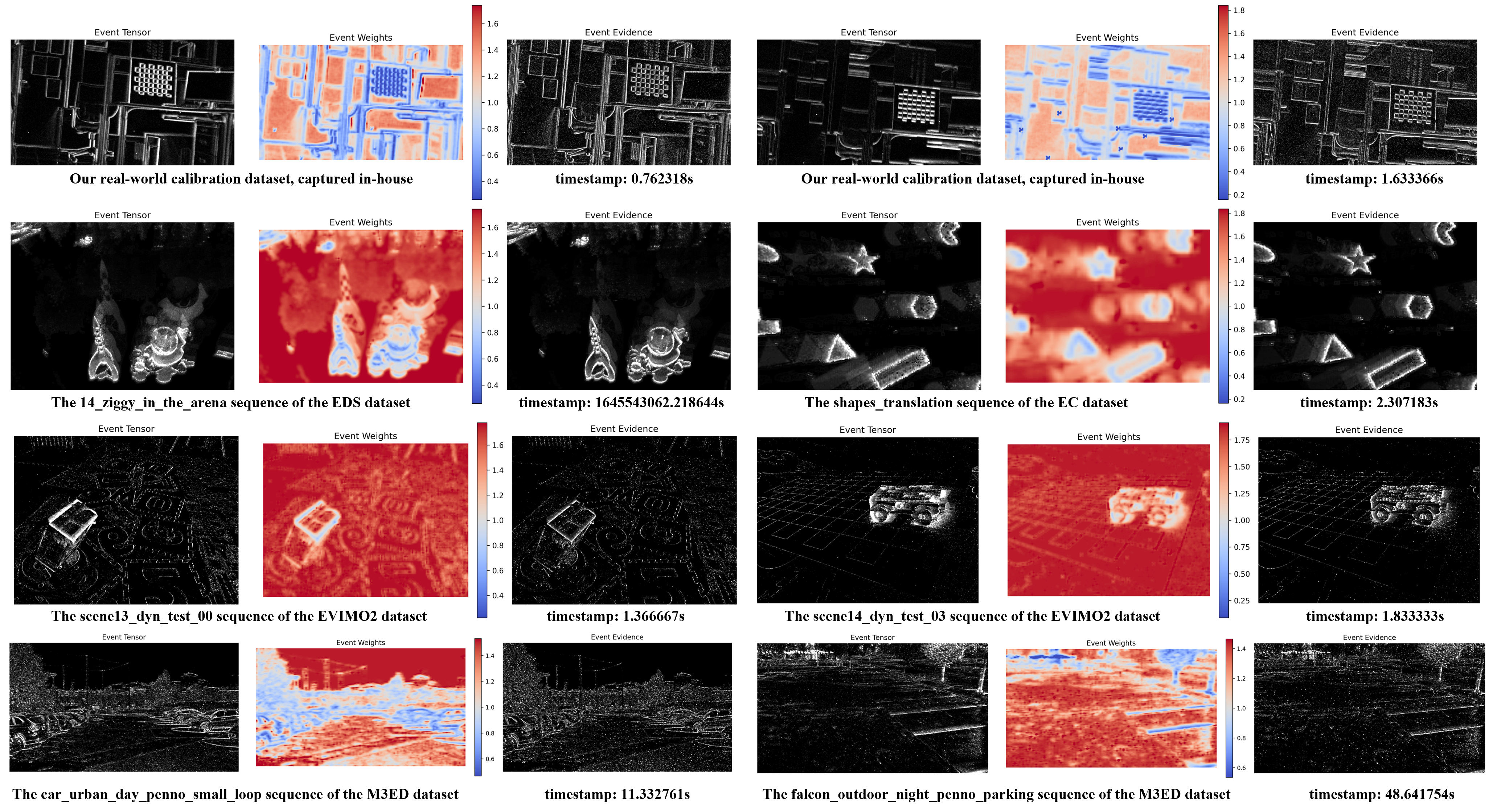}
\caption{TEEM visualizations on real event recordings across datasets and timestamps. Each example shows the input event tensor, the learned event-weight map, and the modulated event evidence. Color bars indicate the weight values.}
\label{fig:teem_visualization}
\end{figure*}

Fig.~\ref{fig:teem_visualization} presents event tensors, TEEM weight maps, and modulated evidence from the in-house calibration sequence and real EC, EDS, EVIMO2, and M3ED recordings. Without edge masks or event-importance labels, TEEM learns spatially structured weights that follow checkerboard boundaries, scene contours, and moving-object outlines even where accumulated events leave motion trails. The modulated evidence retains recognizable geometric structure rather than uniformly amplifying event activity. This task-driven selectivity provides a concrete explanation for the ETAP+TEEM gains over ETAP on several datasets.

\begin{figure}[!t]
\centering
\includegraphics[width=\columnwidth]{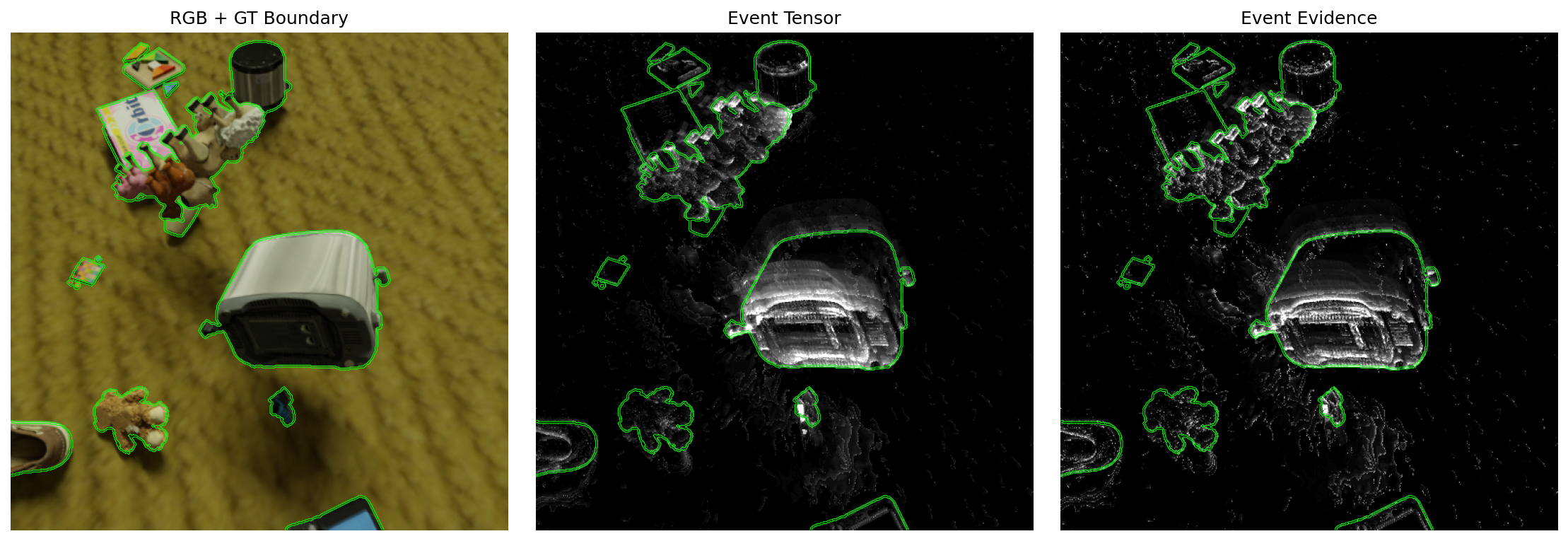}
\caption{TEEM evidence at visible object boundaries on EventKubric. Segmentation-derived ground-truth contours (green) are overlaid on the aligned RGB frame, original event tensor, and TEEM-modulated event evidence.}
\label{fig:teem_boundary}
\end{figure}

To test whether the visual differences reflect better alignment with object contours, we compare the original and modulated event evidence against visible object boundaries from aligned EventKubric RGB and segmentation data (Fig.~\ref{fig:teem_boundary}). Across 10 test sequences (240 frames), TEEM increases the boundary-hit rate from $55.3\%$ to $66.6\%$ and reduces the mean boundary-response width from $4.013$ to $3.359$ pixels ($16.3\%$). These contour-based measurements show that TEEM produces more compact and better-aligned event responses at visible object boundaries.

\subsubsection{Contribution of PTP}
ETAP+PTP improves AJ on both TAP datasets and FA on EC, EDS, and M3ED under both training allocations (Table~\ref{tab:ablation_all}). Unlike TEEM, PTP has no trainable weights; these consistent gains support retaining geometric constraints across window boundaries rather than relying on coordinate initialization alone.

\subsubsection{Combined effects and inference cost}
With the expanded training split, SEPT exceeds either single-module variant in AJ on EventKubric and EVIMO2 and in FA and EFA on EDS. It also improves the primary tracking metric over the retrained ETAP baseline on all five datasets. TEEM and PTP act at complementary stages: TEEM selects event evidence before encoding, whereas PTP carries and refines the resulting trajectory across windows. PTP alone retains the highest EC FA and EFA, showing that the benefit of their combination depends on the evaluation domain.

\begin{figure}[!t]
\centering
\includegraphics[width=0.9\columnwidth]{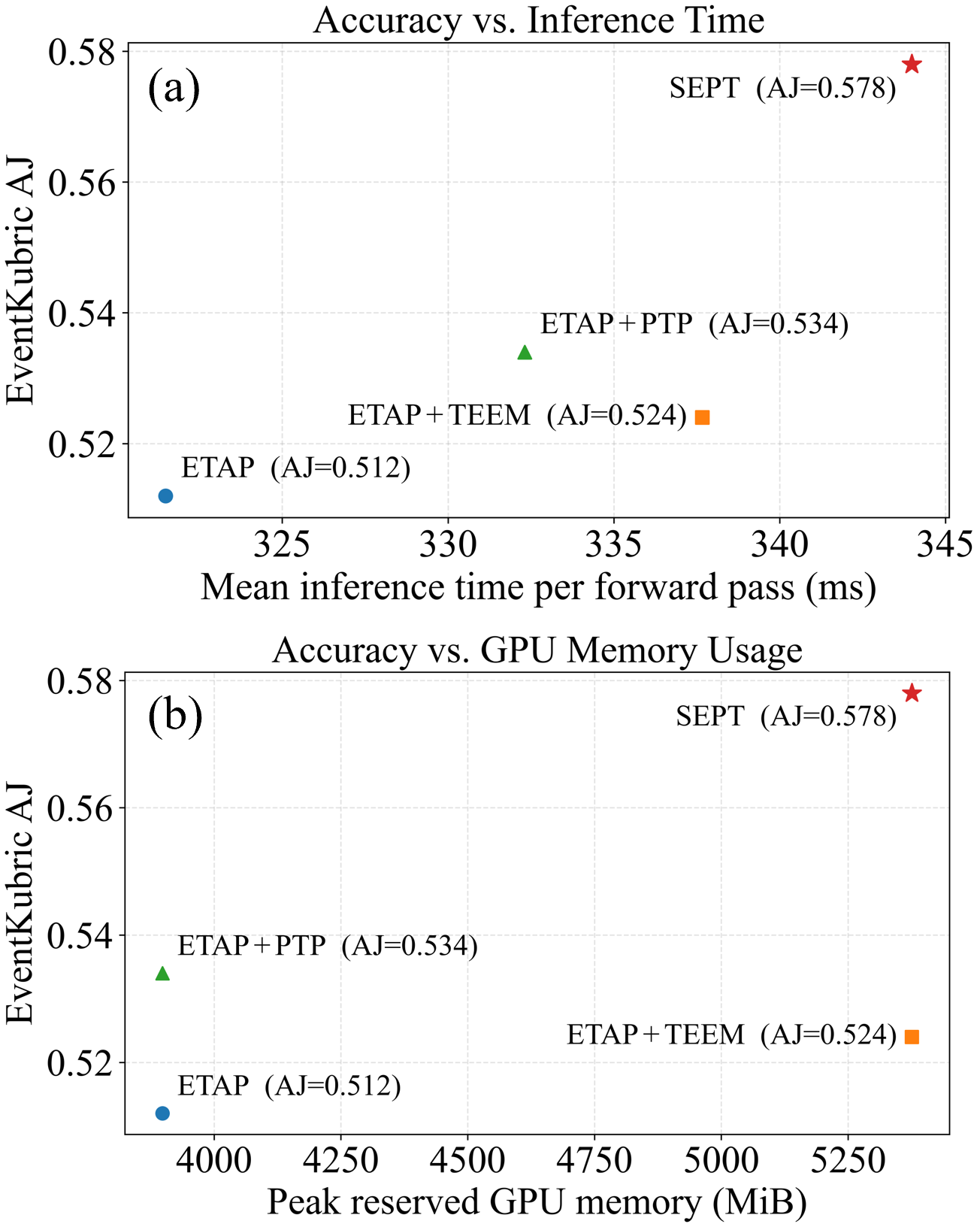}
\caption{EventKubric AJ versus (a) mean inference time per forward pass and (b) peak reserved GPU memory for ETAP and the ablated variants. Peak allocated memory changes little across variants and is not plotted. Results use the expanded training split.}
\label{fig:ablation_efficiency}
\end{figure}

Fig.~\ref{fig:ablation_efficiency} shows the accuracy--resource trade-off under the expanded training split. SEPT achieves the highest EventKubric AJ but also has the longest inference time and ties ETAP+TEEM for the highest peak reserved GPU memory. The memory increase occurs only when TEEM is enabled; ETAP+PTP matches ETAP's peak reserved GPU memory while improving AJ with little added time. PTP therefore offers the most favorable accuracy--efficiency balance among the single-module variants and points to compact trajectory priors as a promising design direction.

\subsubsection{Sensitivity to training-data allocation}
On the shared 80:15:5 validation set, the validation-selected checkpoints yield mean visible-point endpoint error (EPE) values of $6.3902$ for SEPT and $6.6403$ for ETAP+PTP; lower is better. On the EventKubric and EVIMO2 tests under the same 80:15:5 allocation, ETAP+PTP nevertheless has higher AJ than SEPT.

We attribute SEPT's weaker test AJ despite its lower validation EPE to overfitting of the learnable TEEM front end under the smaller synthetic training allocation. PTP provides a strong coordinate-space prior without additional learned weights, whereas the full model needs broader training coverage for TEEM to improve test-time correspondence.

With the 90:5:5 allocation, SEPT overtakes ETAP+PTP in AJ on both TAP test sets and delivers gains over ETAP across all five benchmarks. The additional synthetic examples allow the learned event weighting and the persistent trajectory prior to work together more effectively.

\section{Discussion}
Event cameras respond to brightness changes rather than static appearance. In a static scene with stable illumination, few scene events are triggered, and the remaining event stream may be dominated by sensor noise. A TAP method then receives sparse observations with a low signal-to-noise ratio and few stable features to match. This limitation follows from the sensing principle itself and cannot always be addressed by improving the event-only tracker. A promising direction is multimodal fusion: visible-light images can provide appearance cues when the scene is static, while inertial measurements can supply complementary motion information.

A second challenge is balancing accuracy against processing cost. SEPT attains the highest AJ on both TAP benchmarks but has higher peak reserved GPU memory and longer inference time than ETAP. By contrast, PTP alone improves accuracy with little overhead and without increasing peak reserved GPU memory. Future trackers should retain such compact geometric constraints while reducing the cost of learned event modulation, especially for low-latency applications.

\section{Conclusion}
Stable long-term event-based point tracking requires reliable correspondences from sparse, motion-dependent event activity and trajectory continuity across finite windows. SEPT addresses these complementary challenges with TEEM, which learns task-driven event weights before feature extraction, and PTP, which carries a parameter-free, Schur-marginalized geometric prior through overlapping windows. Trained exclusively on synthetic EventKubric data, SEPT improves TAP AJ on EventKubric and EVIMO2 and feature-tracking FA and EFA on EC, EDS, and M3ED over our retrained ETAP baseline; on EC and EDS, it also surpasses several trackers using both events and frames. Ablations show that PTP provides consistent gains at modest cost and that broader synthetic training improves the full model, which incurs a higher inference cost. These findings support combining selective event evidence with explicit geometric memory. Future work should reduce event-modulation cost and supply complementary cues when event activity is sparse.

\bibliographystyle{IEEEtran}
\bibliography{references}
\end{document}